\documentclass[runningheads]{llncs}
\usepackage[utf8]{inputenc}

\usepackage[T1]{fontenc}
\usepackage{graphicx}
\usepackage{subcaption}
\usepackage[most]{tcolorbox} 
\usepackage{enumitem}        
\usepackage{booktabs}
\usepackage{lipsum}
\usepackage{multirow}
\usepackage{tabularx, booktabs}
\usepackage[colorlinks=true, urlcolor=blue]{hyperref}

\begin{document}
\title{Assessing On-the-Ground Disaster Impact Using Online Data Sources}
%
%
\author{Saketh Vishnubhatla\inst{1} \and
Ujun Jeong\inst{1} \and
Bohan Jiang\inst{1} \and
Paras Sheth\inst{1} \and
Zhen Tan\inst{1} \and \\
Adrienne Raglin \inst{2} \and
Huan Liu\inst{1}}
\authorrunning{S. Vishnubhatla et al.}  
%
\institute{Arizona State University, Tempe, USA \\
\email{\{svishnu6,ujeong1,bjiang14,psheth5,ztan36,huanliu\}@asu.edu}
\and
DEVCOM Army Research Lab, USA\\
\email{adrienne.raglin2.civ@army.mil}}
\maketitle              
\begin{abstract}
Assessing the impact of a disaster in terms of asset losses and human casualties is essential for preparing effective response plans. Traditional methods include offline assessments conducted on the ground, where volunteers and first responders work together to collect the estimate of losses through windshield surveys or on-ground inspection. However, these methods have a time delay and are prone to different biases. Recently, various online data sources, including social media, news reports, aerial imagery, and satellite data, have been utilized to evaluate the impact of disasters. Online data sources provide real-time data streams for estimating the offline impact. Limited research exists on how different online sources help estimate disaster impact at a given administrative unit. In our work, we curate a comprehensive dataset by collecting data from multiple online sources for a few billion-dollar disasters at the county level. We also analyze how online estimates compare with traditional offline-based impact estimates for the disaster. Our findings provide insight into how different sources can provide complementary information to assess the disaster.
\end{abstract}
\section{Introduction}

The disaster management cycle includes mitigation and preparedness phases before the disaster, followed by response and recovery phases during and after the disaster \cite{sawalha2020contemporary}. One of the crucial aspects of responding to and recovering from a disaster is to assess the impact of a disaster. The impact of any disaster can be assessed in terms of losses to assets and human lives. Initial loss estimates are often made through windshield surveys and in-situ inspections. Further, government agencies work offline (on-ground) in collaboration with multiple local volunteers and first response teams to collect loss estimates. 

While offline statistics are often considered reliable, they may be conservative and subject to various biases. Some common biases in disaster loss databases include accounting, geographical, and temporal biases \cite{gall2009when,zhou2025knowing}. Moreover, there is a time lag in obtaining the loss estimates from these loss databases. In this regard, online data sources like news, social media, aerial imagery, and satellite imagery provide a near-real-time data stream \cite{kryvasheyeu2016rapid,yamazaki2007remote,okada2021potential}. News media offers a global context of the disaster as it unfolds. Social media offers crowdsourced information that can help gauge the impact of a disaster. Aerial imagery can provide an isolated view of different regions, while satellite imagery from remote sensing provides a comprehensive view of the land cover changes during a disaster. Thus, online sources can help estimate the impact of a disaster in real time.


However, the value of online data is enhanced only if we aggregate all sources at a well-defined administrative unit to estimate the offline impact within the unit. This process requires geolocating multiple sources of information: news articles, social media posts, and satellite imagery. 

In this work, we study the utility of different online sources in estimating the offline impact of a disaster. We create a dataset comprising news articles, social media posts, and land cover transition features from satellite imagery. Our dataset is aggregated at a county level, allowing us to compare the online estimates with recorded offline county-level losses.

\section{Related Work}

\subsection{Offline Assessment} 
Disaster impact loss estimates are traditionally obtained through on-the-ground data collection. For instance, FEMA follows a preliminary disaster assessment \cite{fema2020pda}, where the disaster-affected areas are inspected on the ground to get a quick estimate of the losses. This assessment can later be used to call upon a presidential declaration by the governor. There exist many disaster loss databases, such as SHELDUS\footnote{\url{https://cemhs.asu.edu/sheldus}}, NOAA\footnote{\url{https://www.ncdc.noaa.gov/stormevents/}}, and NHC\footnote{\url{https://www.nhc.noaa.gov/data/}}. SHELDUS is an aggregator for disaster losses ranging from 1960 to 2023 for 18 natural hazards types. SHELDUS data can be queried at a county level. With NOAA storms and unusual weather phenomena, the data provided is at a forecast zone level, and the data is updated every month. The NHC data archive is a hurricane-specific archive that provides information about hurricane tracks, wind intensities, etc., without information about the losses.

\subsection{Online Assessment}
\textbf{Media-Based Estimation.}
Previous works have explored using social media to assess disasters. Datasets such as CrisisMMD \cite{alam2018crisismmd}, DMD \cite{mouzannar2018damage}, and CrisisBench \cite{alam2021crisisbench} compile microblogging posts that include both images and captions. These works aim to classify each post by severity level or humanitarian category. Recent studies have also explored using news articles to assess flood risk \cite{fu2022extracting,fu2025creating}.\\ \\
\textbf{Satellite and Aerial Imagery.}
Open-access products such as MODIS, Landsat, and Sentinel offer information on vegetation indices and land cover transitions, varying in spatial resolution and revisit frequency. ResueNet \cite{rahnemoonfar2023rescuenet} is a machine learning model to assess building damages over high-resolution imagery by looking at the pre-disaster and post-disaster images. Another key resource is Google’s Dynamic World \cite{brown2022dynamic}, which generates classified land cover maps from Sentinel-2 imagery and is widely used to monitor seasonal and disaster-induced land changes. The use of drones and aerial surveys for disaster assessment is rapidly expanding. In the aftermath of hurricanes, drones are deployed to assist in locating survivors in inaccessible or remote areas \cite{greenwood2020flying}. Another study focuses on assessing the damage caused by landslides through aerial imagery \cite{chang2020uav}.

\section{Data Collection}
\begin{figure}[htbp]
  \centering
  \begin{subfigure}[b]{0.66\columnwidth}
    \centering
    \includegraphics[width=\linewidth]{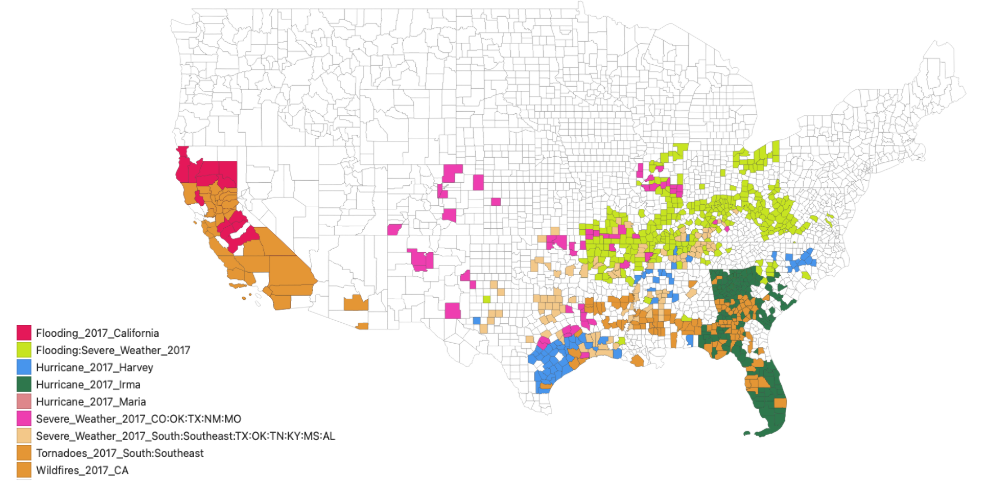}
    \caption{}
    \label{fig:counties_of_interest}
  \end{subfigure}
  \hfill
  \begin{subfigure}[b]{0.33\columnwidth}
    \centering
    \includegraphics[width=\linewidth]{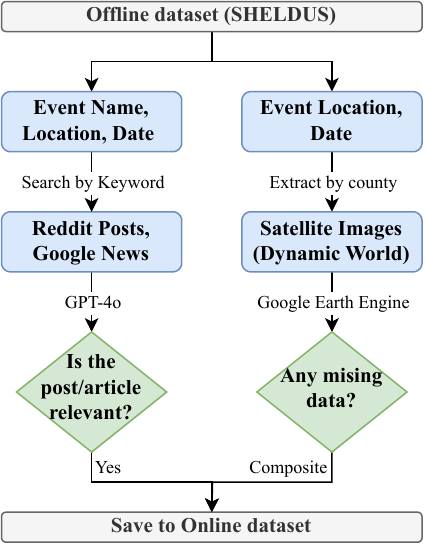}
    \caption{}
    \label{fig:reddit_collection}
  \end{subfigure}
  \caption{%
    (a) Collected counties and events from SHELDUS. %
    (b) Data collection pipeline for online sources: news, social media data, and satellite images.
  }
  \label{fig:combined}
\end{figure}

Our analysis focuses on disasters with at least a billion dollars in damages in the U.S. As shown in Figure~\ref{fig:counties_of_interest}, we collect nine billion-dollar events\footnotemark\ from 2017 using the SHELDUS database. We chose the county as the administrative unit of analysis. An overview of our data collection process is shown in Figure~\ref{fig:reddit_collection}. Utilizing the disaster name and location information, we gathered data from various online sources for 1,097 county-event pairs. 

\footnotetext{
\textbf{Event Index Used In This Paper:}
\textbf{1:} Flooding 2017 Midwest--Central,
\textbf{2:} Flooding 2017 California,
\textbf{3:} Hurricane 2017 Harvey,
\textbf{4:} Hurricane 2017 Irma,
\textbf{5:} Hurricane 2017 Maria,
\textbf{6:} Severe Weather 2017 CO--OK--TX--NM--MO,
\textbf{7:} Severe Weather 2017 South--Southeast,
\textbf{8:} Tornadoes 2017 South--Southeast,
\textbf{9:} Wildfires 2017 CA.
}

\subsection{Offline Data Sources}
We use SHELDUS as the source for disaster losses. SHELDUS provides user-friendly data downloads, unlike other counterparts like the NOAA database, which cannot be directly aggregated at a county level without additional preprocessing. SHELDUS provides losses on four metrics:  property and crop damages in dollars and injury and fatality counts for every county for an event.

\subsection{Online Data  Sources}
We focus on Reddit for social media, while for news media, we scrape articles from Google News using GNews API \cite{rahaani2023gnews}. For satellite imagery, Dynamic World gathers land cover maps for the region of interest. Our key contribution is that we aggregate all the data at the county level.

\smallskip

\noindent\textbf{Social Media Data from Reddit:}
Conventional social media geolocation-based methods use the user's profile and posts to assign a possible location. Some other approaches use geotagging metadata associated with the posts, though most of the data on conventional social media platforms will not have such tags. In our work, we instead geotag subreddits to their most likely counties. ``r/LocationSubreddits'' provides a list of subreddits for different cities/counties across the U.S. We use Nominatim \cite{nominatim}, an open street-map based library to map different subreddit names (e.g., ``r/fortlauderale'') to their associated counties. From each subreddit for a county, we use a set of keywords, e.g., ``storm'' from ``r/Austin'', to retrieve relevant articles for a disaster. In addition, we use GPT-4o-mini to filter relevant posts based on disaster summaries provided in the prompt. We use Reddit pushshift archive \cite{baumgartner2020pushshift} to retrieve data from Reddit. In total, we map around 3000 posts from over 54 subreddits.

\smallskip

\noindent\textbf{News Media Data from Google News:}
We use GNews API to crawl for relevant news articles. To geolocate news articles, we append disaster-specific keywords with county names (e.g., ``flood Merced'') to search for news articles. Later, we use GPT-4o-mini to filter relevant posts, mapping nearly 7000 entries for different county-event pairs. 

\smallskip

\noindent\textbf{Satellite Data from Dynamic World:}
Dynamic World provides maps with nine land cover classes. The spatial resolution is at a 10m level with a refresh frequency of 2-5 days. Each pixel on the Sentinel-2 satellite image is marked into one of the land cover classes. One common issue with remote sensing products is handling missing data from cloud covers or sensor issues. We use Google Earth Engine \cite{gorelick2017google}, which allows compositing. With compositing, multiple images over a time window are patched to construct a complete image. We create two composite images for each county, before and after the disaster. Later, 81 landcover transition features are computed, comprising area change in meters squared from a source land cover class (e.g., tree) to a target land cover class (e.g., water) within the county.


\section{Experiments and Results}

\subsection{Experimental Setup}

Our task formulation involves estimating various categories of disaster severity at the county level for a given disaster. Specifically, for each of the 1,097 county-disaster combinations in the dataset, the goal is to predict severity levels for property damage, crop damage, injuries, and fatalities. For property damage, severity is classified into three levels: low (less than \$10,000), medium (between \$10,000 and \$1 million), and high (greater than \$1 million). Similarly, crop damage is categorized as low (under \$1,000), medium (between \$1,000 and \$100,000), and high (above \$100,000). Injury and fatality estimation tasks are treated as binary classification tasks, with ``yes'' and ``no'' indicating the presence or absence. These labels are derived from the SHELDUS and serve as ground truth for evaluating task performance. We independently evaluate remote sensing-based and media-based estimations, reflecting their distinct modeling paradigms, supervised learning in the former and LLM-based zero-shot learning in the latter.

\medskip

\noindent\textbf{Remote-sensing-based estimation:} We use land cover transition before and after the disaster (e.g., area change (in meters squared) over a county from built-up to water), resulting in 81 input features per county. We use different buffer windows (15, 30, 45, and 60 days) to generate the composites. Different models used are shown in Table~\ref{tab:model-configs}.
\vspace{-2em}
\begin{table}[ht]
\centering
\caption{Model configurations and hyperparameters used for severity estimation.}
\begin{tabularx}{\linewidth}{@{}lX@{}}
\toprule
\textbf{Model} & \textbf{Configuration} \\
\midrule
Logistic Regression & L2 penalty, L-BFGS optimizer, max\_iter = 300 \\
Random Forest       & 100 trees, max depth = 10 \\
XGBoost             & 100 rounds, learning rate = 0.1, L1/L2 regularization $\lambda = \alpha = 0.1$ \\
MLP (1 layer)       & 128 units \\
MLP (2 layers)      & 128, 64 units \\
MLP (5 layers)      & 256, 128, 64, 32, 16 units \\
\bottomrule
\end{tabularx}
\label{tab:model-configs}
\end{table}

\vspace{-1em}

\begin{figure}[!htbp]
  \centering
  \begin{subfigure}[b]{\linewidth}
    \centering
    \includegraphics[width=0.84\linewidth]{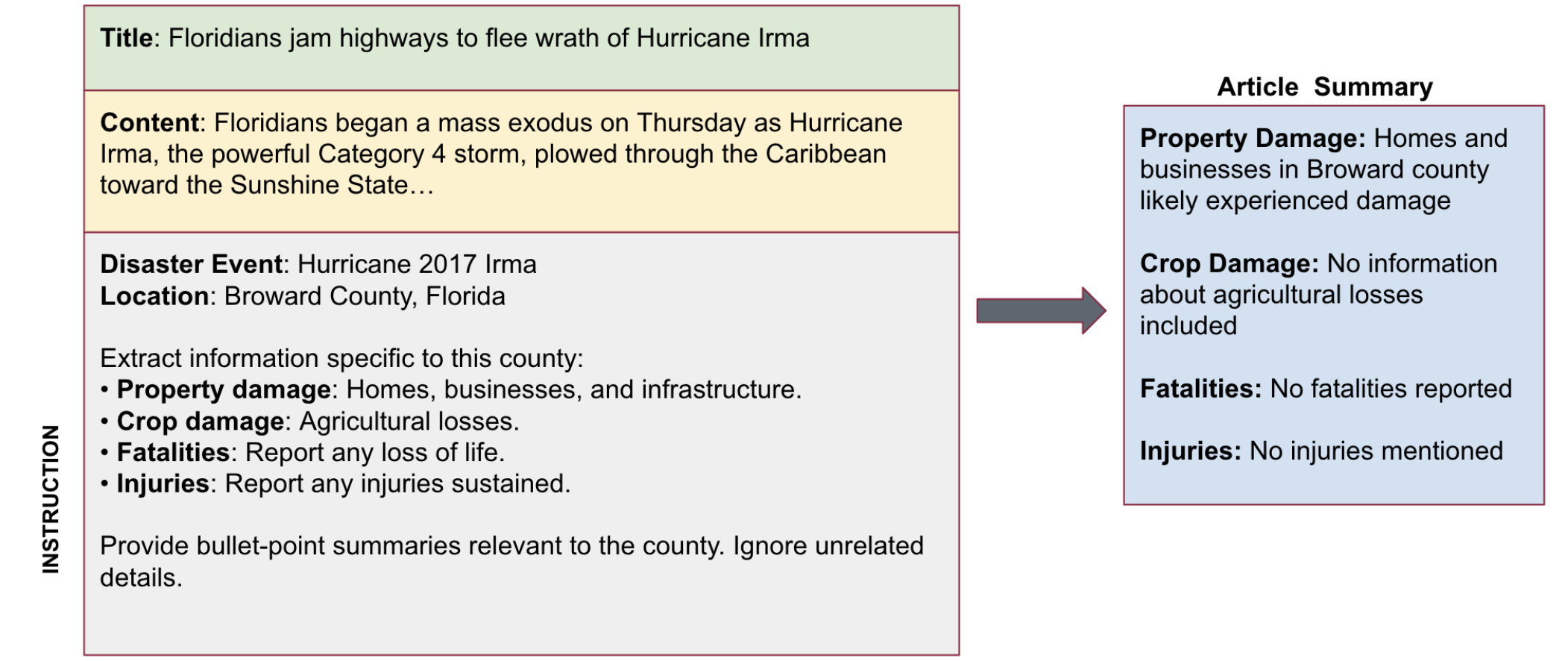}
    \label{fig:phase1}
  \end{subfigure}
  \vspace{0.2cm}
  \begin{subfigure}[b]{\linewidth}
    \centering
    \includegraphics[width=0.8\linewidth]{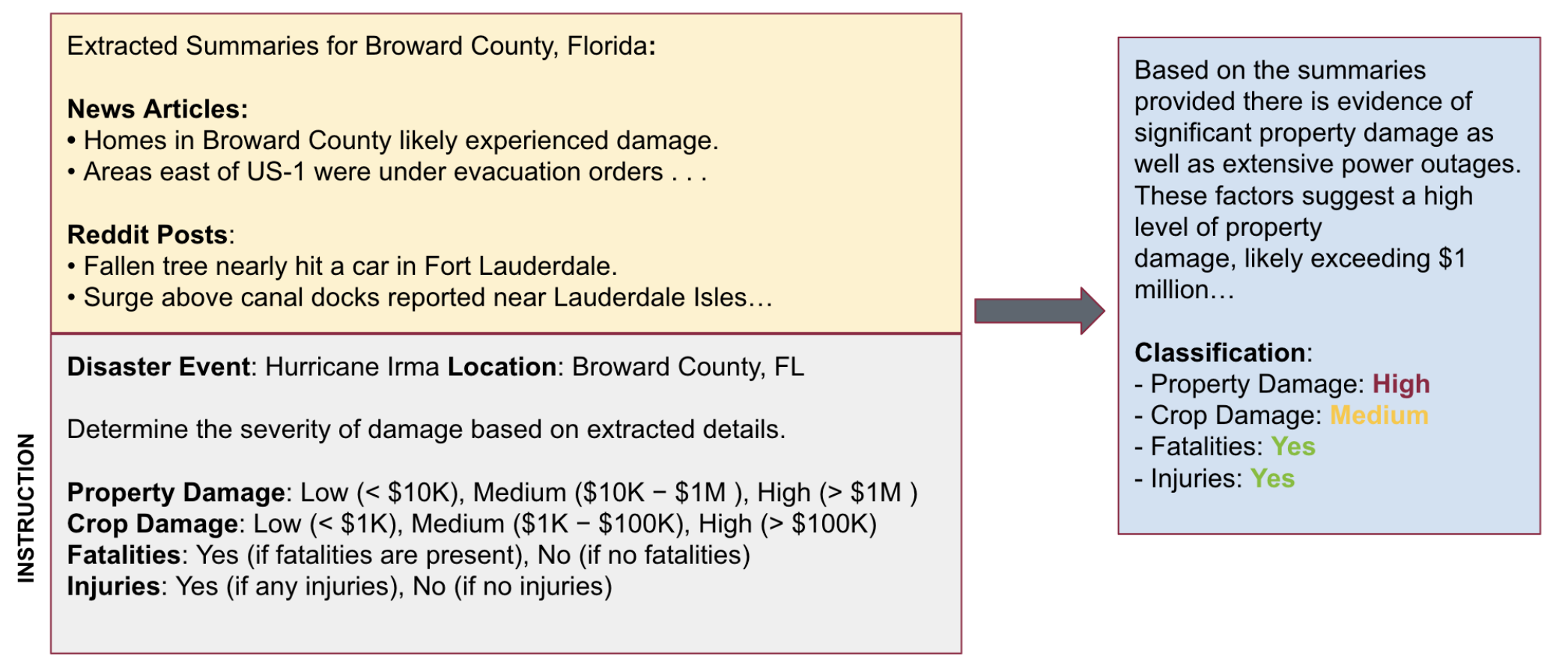}
    \label{fig:phase2}
  \end{subfigure}
  \caption{Two-phase pipeline for disaster assessment using media data. On the top we see a sample prompt for the summary extraction step; the bottom shows the severity estimation using county-level summaries extracted in phase 1.}
  \label{fig:two_phase}
\end{figure}

\noindent\textbf{Social and news media-based estimation:} We follow a two-stage prompting approach (see Figure~\ref{fig:two_phase}) using large language models (LLMs) such as GPT-4o, GPT-4o-mini, and LLaMA3-8B-Instruct. In the first stage, these models extract structured summaries of disaster impact, covering property and crop damage, injuries, fatalities, and relief, from county-level news articles and Reddit posts. In the second stage, classification prompts are used to predict severity labels from these summaries.

\begin{table*}[!htbp]
\centering
\caption{Geolocation and relevance accuracy (\%) for each disaster event.}
\resizebox{\textwidth}{!}{
\begin{tabular}{lccccccccc}
\toprule
\textbf{Metric} & \textbf{Event 1} & \textbf{Event 2} & \textbf{Event 3} & \textbf{Event 4} & \textbf{Event 5} & \textbf{Event 6} & \textbf{Event 7} & \textbf{Event 8} & \textbf{Event 9} \\
\midrule
Geo. Acc. (\%) & 8.3 & 91.7 & 25.0 & 25.0 & 75.0 & 0.0 & 16.7 & 75.0 & 83.3 \\
Rel. Acc. (\%)   & 50.0 & 91.7 & 100.0 & 91.7 & 91.7 & 58.3 & 50.0 & 75.0 & 100.0 \\
\bottomrule
\end{tabular}
}
\label{tab:geo_relevance_accuracy_indexed}
\end{table*}

Despite applying a filtering process, the collection of news articles remains susceptible to noise. To better understand data quality, we sample 12 articles for each event type, spanning various counties and events, and conduct manual verification for relevance and geolocation accuracy (see Table~\ref{tab:geo_relevance_accuracy_indexed}). Our analysis reveals higher geolocation accuracy for events such as California Flooding (Event 2), Hurricane Maria (5), Southeast Tornadoes (8), and California Wildfire (9). In contrast, for Hurricane Harvey (3) and Irma (4), the larger volume of available content enables the LLM-based two-phase pipeline to extract more meaningful information. Therefore, we use California Flooding (2), Harvey (3), Irma (4), Southeast Tornadoes (8), and California Wildfires (9) for our analysis with social and news media estimation. We leave out Hurricane Maria (5) owing to a lack of Reddit data.

\vspace{-3em}

\subsection{Performance Comparison Across Information Sources}

\begin{figure}[!htbp]
  \centering
  \includegraphics[width=0.95\textwidth]{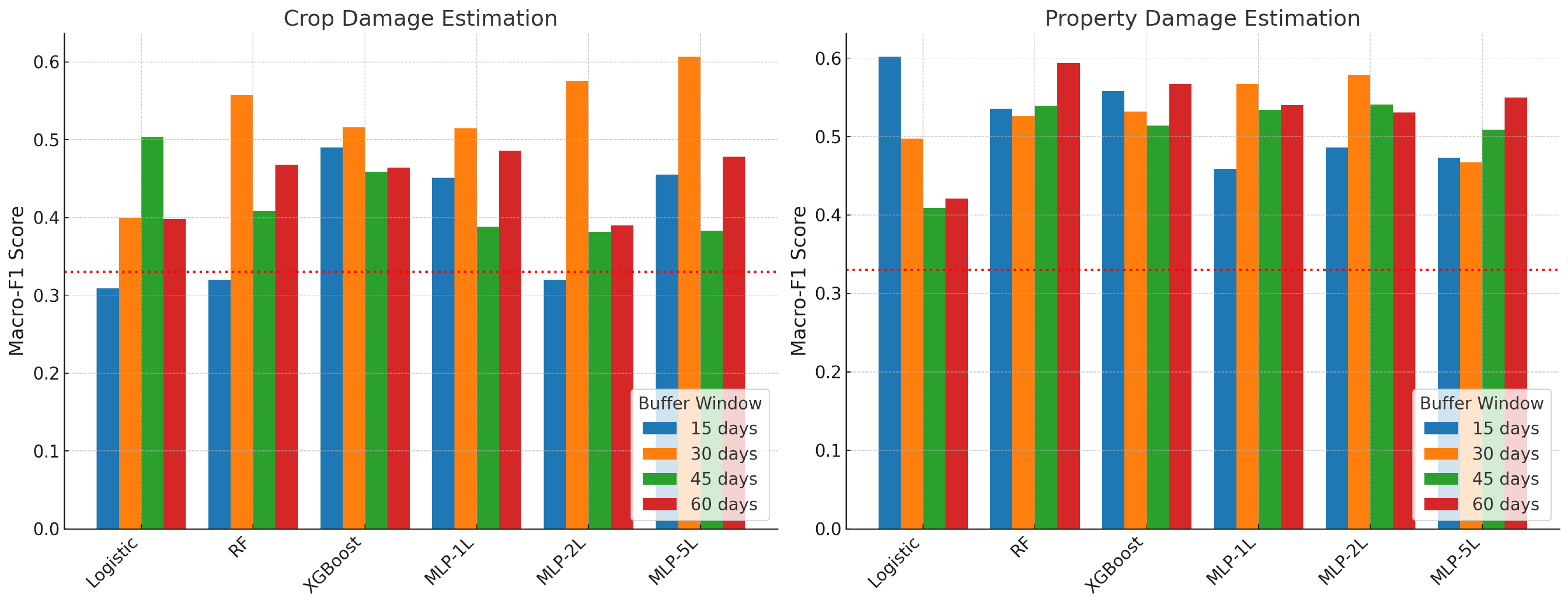} 
  \caption{Performance with remote sensing data. The dotted red line indicates expected performance from a random baseline classifier.}
  \label{fig:rem_sens_performance}
\end{figure}

Satellite imagery captures physical land cover changes, and is inherently suited for measuring changes in the built-up and agricultural environment but is unable to detect human-centric outcomes like injuries or fatalities. Therefore, remote sensing is used exclusively for property and crop damage estimation in our setup. In contrast, digital and social media sources (news and Reddit) are applied across all four tasks, including injuries and fatality detection. 

We see the performance for remote sensing methods in Figure~\ref{fig:rem_sens_performance}. We find the best performances with a buffer window size of 30 days for crop damage estimation for different models. Longer window sizes may disturb the seasonal variations needed to accurately assess crop damage. However, when it comes to property damage, we observe that in most cases, the performance improves as the buffer window increases. This can be understood as property damage changes are more permanent and show up even as the window size increases.

With news and social media-based estimation, their predictive utility is most evident over the fatality detection task. Though the informativeness of these sources for crop damage, property damage, and injury estimation remains limited. The results can be seen in Table~\ref{tab:macro_f1_news_reddit_only}. 

\begin{table*}[ht]
\centering
\caption{Macro-F1 scores across five events by model, source, and category. Highlighted performances of above 0.5 property/crop damage estimation, and above 0.65 for injury/fatality estimation tasks.}
\resizebox{\textwidth}{!}{
\begin{tabular}{lllccccc}
\toprule[1.5pt]
\textbf{Model} & \textbf{Source} & \textbf{Category} & \textbf{Event 2} & \textbf{Event 3} & \textbf{Event 4} & \textbf{Event 8} & \textbf{Event 9} \\
\midrule

\multirow{8}{*}{\textbf{GPT-4o}}
& \multirow{4}{*}{News}
  & Property   & 0.306 & \textbf{0.543} & 0.264 & 0.241 & 0.151 \\
& & Crop       & 0.130 & 0.310 & 0.211 & 0.223 & 0.061 \\
& & Injuries   & 0.408 & 0.524 & 0.431 & 0.609 & 0.479 \\
& & Fatalities & \textbf{0.672} & \textbf{0.692} & 0.625 & 0.648 & \textbf{0.670} \\
\cmidrule(lr){2-8}
& \multirow{4}{*}{Reddit}
  & Property   & 0.222 & 0.389 & 0.246 & 0.182 & 0.167 \\
& & Crop       & 0.311 & 0.222 & 0.370 & 0.074 & 0.000 \\
& & Injuries   & 0.385 & 0.286 & 0.077 & 0.273 & 0.143 \\
& & Fatalities & 0.429 & 0.429 & 0.625 & 0.467 & 0.333 \\
\midrule

\multirow{8}{*}{\textbf{GPT-4o-mini}}
& \multirow{4}{*}{News}
  & Property   & 0.306 & 0.287 & 0.295 & 0.178 & 0.087 \\
& & Crop       & 0.101 & 0.194 & 0.135 & 0.253 & 0.087 \\
& & Injuries   & 0.383 & 0.392 & 0.387 & 0.453 & 0.487 \\
& & Fatalities & \textbf{0.731} & \textbf{0.740} & \textbf{0.660} & 0.570 & \textbf{0.687} \\
\cmidrule(lr){2-8}
& \multirow{4}{*}{Reddit}
  & Property   & 0.376 & 0.319 & 0.333 & 0.182 & 0.167 \\
& & Crop       & 0.281 & 0.311 & 0.222 & 0.074 & 0.000 \\
& & Injuries   & 0.467 & 0.351 & 0.286 & 0.111 & 0.143 \\
& & Fatalities & 0.429 & 0.537 & 0.429 & 0.385 & 0.486 \\
\midrule

\multirow{8}{*}{\textbf{LLaMA3-8B}}
& \multirow{4}{*}{News}
  & Property   & 0.240 & 0.277 & 0.169 & 0.270 & 0.000 \\
& & Crop       & 0.133 & 0.139 & 0.133 & 0.259 & 0.000 \\
& & Injuries   & 0.300 & 0.238 & 0.240 & 0.519 & 0.000 \\
& & Fatalities & 0.566 & 0.295 & 0.268 & 0.509 & 0.111 \\
\cmidrule(lr){2-8}
& \multirow{4}{*}{Reddit}
  & Property   & 0.111 & 0.000 & 0.000 & 0.378 & 0.000 \\
& & Crop       & 0.083 & 0.133 & 0.083 & 0.133 & 0.000 \\
& & Injuries   & 0.273 & 0.000 & 0.077 & 0.300 & 0.000 \\
& & Fatalities & 0.300 & 0.200 & 0.100 & 0.417 & 0.000 \\
\bottomrule[1.5pt]
\end{tabular}
}
\label{tab:macro_f1_news_reddit_only}
\end{table*}

\vspace{-0.5em}
\subsection{Perceived Severity vs Recorded Severity}

\begin{figure}[htbp]
  \centering  
  \includegraphics[width=0.55\textwidth]{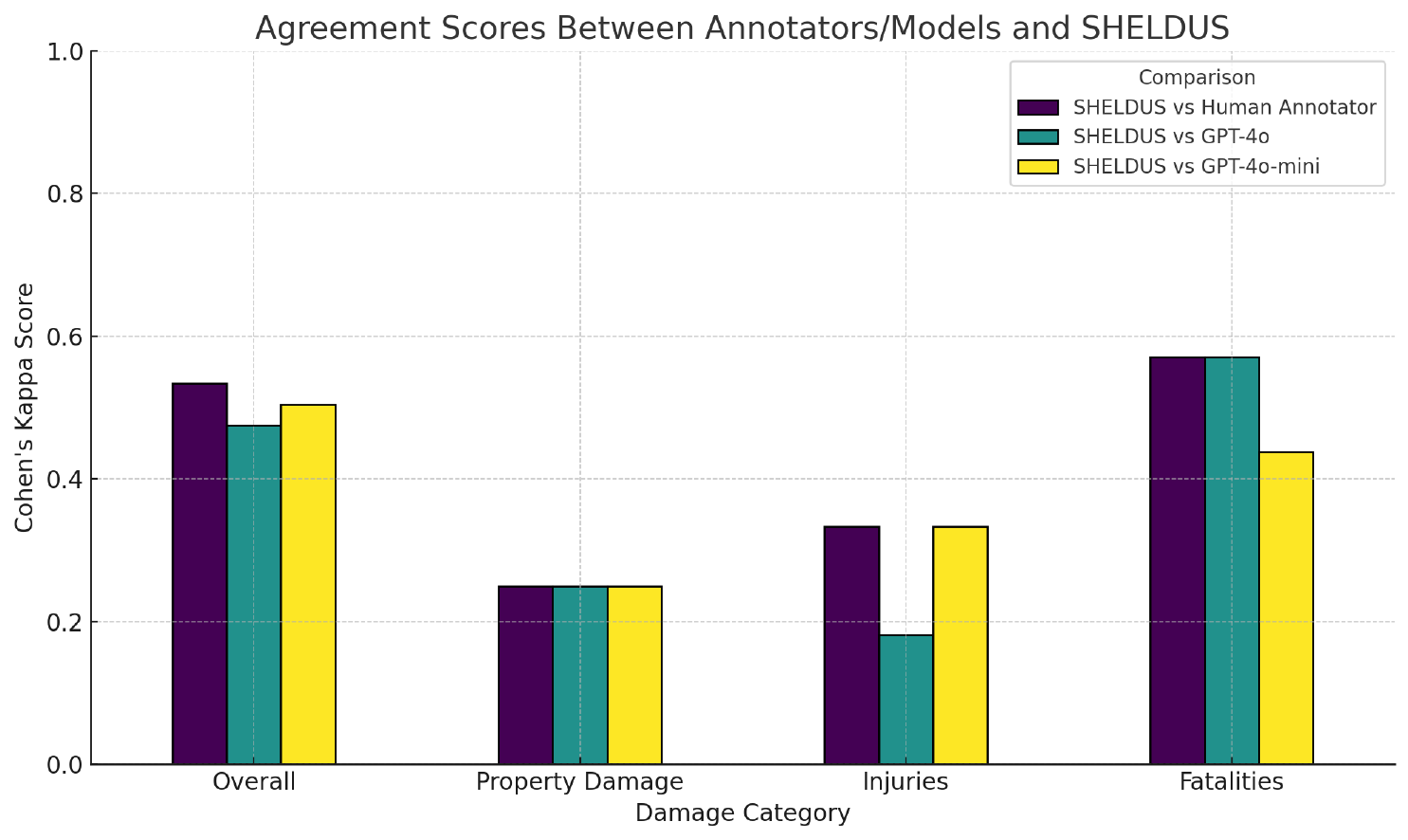}
  \caption{Cohen's Kappa agreement scores between SHELDUS statistics and human annotators or models (GPT-4o, GPT-4o-mini) across damage categories.}
  \label{fig:agreement_scores}
\end{figure}

We selected six counties each from Hurricane Irma and the California Wildfires to assess severity based on available information from the news. Notably, there were discrepancies between human-perceived severity and officially recorded losses from SHELDUS. While SHELDUS offers conservative estimates based on direct damages, Reddit posts and news articles often report deaths, injuries, or losses that may not be directly attributable to the disaster. For example, although fatalities and injuries were reported on social and news media for Orange County, Florida, the offline records do not reflect such impacts. Additionally, geotagging errors occasionally misled the models, contributing to inconsistencies between predictions and ground truth. Across 36 total evaluations (12 counties on 3 tasks: property damage, injuries, fatalities), both LLM outputs and human annotations were analyzed. Cohen's kappa was around 0.5 (see Figure~\ref{fig:agreement_scores}), indicating moderate agreement among annotators. Disagreement is particularly evident in the property damage estimation task. This disagreement is due to the ease of classifying a county in ``low'' or ``high'' severity based on perception, but it is difficult to categorize into the ``medium'' class. For injuries and fatalities, online sources often reported incidents that were either unrecorded offline or ambiguous in terms of their direct connection to the disaster. Although GPT-4o-mini demonstrated slightly better overall performance as seen in Table~\ref{tab:macro_f1_news_reddit_only}, GPT-4o proved to be more reliable when examining the extracted summaries used to arrive at the final estimates.


\section{Discussion}
There is a need for a real-time disaster assessment framework using online data sources. Most of the previous work focused on assessing the impact of events, observing land cover changes, or estimating the sentiment of each article or post. However, none of the work has observed how online sources could be reasonably used to estimate the offline impact of disasters.

Based on our analysis, online sources provide complementary information to assess a disaster. With remote sensing data, we find a moderate performance of property and crop damage estimation. One potential challenge is to mitigate the confounding effect of seasonal changes in land cover and tease out a disaster's causal impact on land cover transitions. News articles have helped estimate fatalities in a county. 

Our analyses show examples of online sources that mention injuries and fatalities, though the losses recorded in SHELDUS were conservative. These findings provide an optimistic outlook that we could leverage multiple online sources to assess a disaster's impact. 

Our study has a few limitations. First, our collection of media data is prone to selection bias, where some counties receive more coverage on social media or have more active local news outlets. Moreover, there is the effect of missing data with satellite imagery, which we try to mitigate by compositing. As a result of selection bias and missing data, some regions remain data ``black spots.''

\section{Conclusion}
In this work, we curated a county-level, multi-source, and multi-modal dataset for disaster assessment\footnote{\url{https://drive.google.com/drive/folders/1yx73x9YxdjNbpD_X1xmCIburvfQ2u_WQ?usp=sharing}}. We systematically studied the performance of utilizing different sources in estimating a disaster's impact. We provided sufficient insights for future research to leverage online sources to assess the effects of a disaster. While our study is limited to a single calendar year, expanding the data curation to multiple years would allow for temporal trend analysis. In addition, using various data sources in combination will also be an interesting direction to pursue. Finally, selecting causally relevant features that drive the classification of a county into a severity level would help build explainable decision-making systems that can be deployed in the real world.


%
%
%
%




\bibliographystyle{unsrt}
\bibliography{references}
\end{document}